\documentclass[letterpaper]{article} 
\usepackage[]{aaai2027}  
\usepackage[hyphens]{url}  
\usepackage{graphicx} 
\usepackage{natbib}  
\usepackage{caption} 
\usepackage{algorithm}
\usepackage{amssymb}
\usepackage{algorithmic}
\usepackage{multirow}      
\usepackage{array}         
\usepackage{colortbl}      
\definecolor{ourrow}{RGB}{230, 240, 255}

\usepackage{newfloat}
\usepackage{listings}
\DeclareCaptionStyle{ruled}{labelfont=normalfont,labelsep=colon,strut=off} 

\usepackage{booktabs}
\definecolor{rowgray}{HTML}{EDEDED}
\definecolor{myblue}{RGB}{155, 0, 0} 
\definecolor{rowpink}{RGB}{255, 230, 235}  
\definecolor{rowours}{HTML}{E8F1FA}
\usepackage{xcolor}
\definecolor{myorange}{RGB}{230, 126, 34}
\definecolor{deepred}{RGB}{139,0,0}
\usepackage{amsmath}

\title{\raisebox{-0.3\height}{\includegraphics[height=1.3cm]{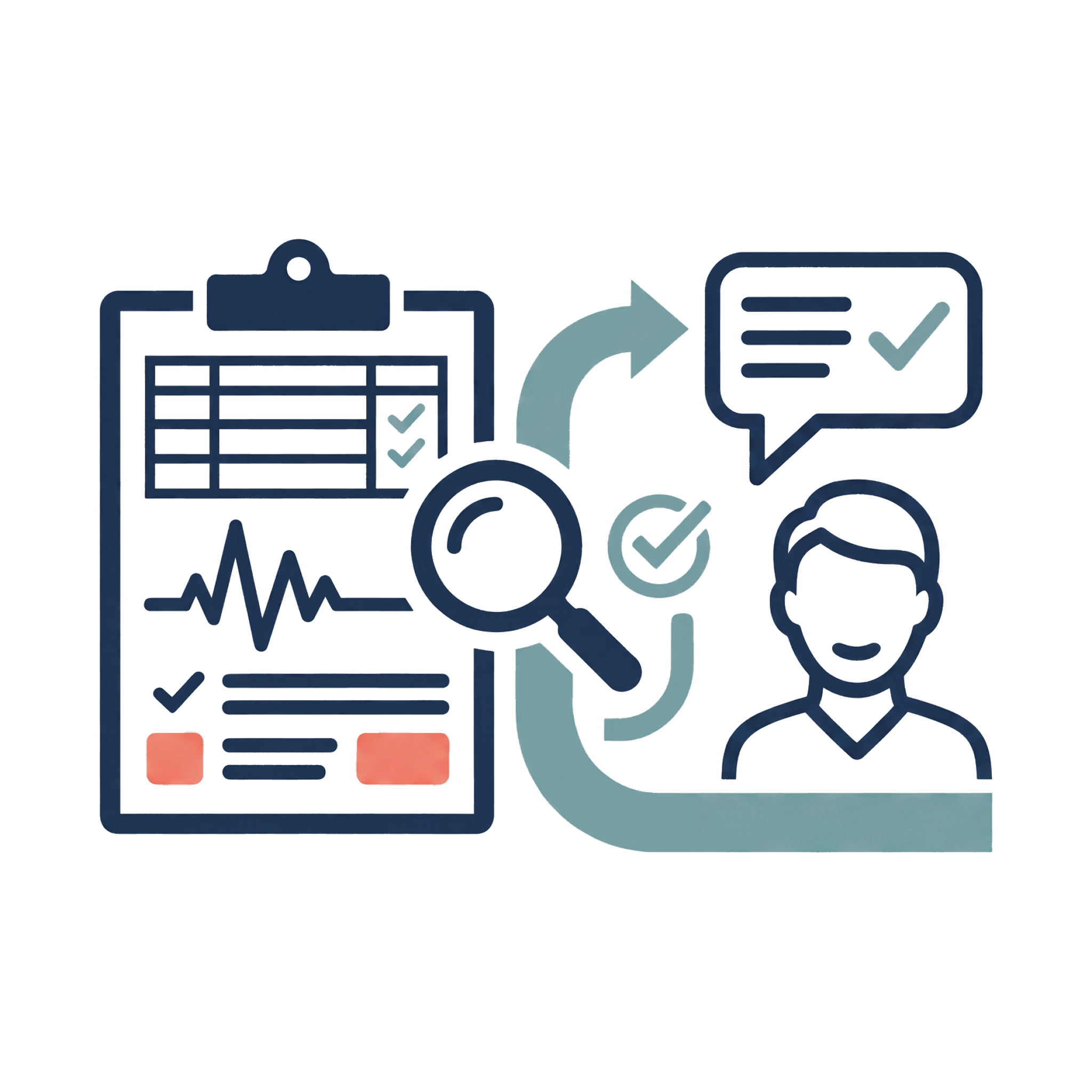}} G-CARL: Grounded Checklist-Aligned Reward Learning for Patient-Oriented Medical Report Interpretation}
\author {
    Shiao Xie\textsuperscript{\rm 1}\equalcontrib,
    Siyu Chen\textsuperscript{\rm 1}\equalcontrib,
    Jianwei Lv\textsuperscript{\rm 1},
    Bo Yuan\textsuperscript{\rm 1},
    Yujin Wang\textsuperscript{\rm 1}\corresponding,
    Xiandong Li\textsuperscript{\rm 1}\corresponding
}
\affiliations {
    \textsuperscript{\rm 1}Baidu Inc., China
}

\nocopyright 
\usepackage{xr}
\begin{document}

\maketitle

\begin{abstract}
Personalized interpretation of medical reports has emerged as an increasingly important need among patients. Addressing this need requires both evidence-grounded medical factuality and context-dependent patient communication, yet existing medical vision-language tasks do not adequately capture these dual requirements.
To bridge this gap, we introduce \emph{Patient-oriented Medical Report Interpretation} (PMRI), a novel open-ended multimodal generation task that requires models to explain medical reports in accurate and accessible language based on a user's query and dialogue history.
These two objectives differ fundamentally in their verifiability, yet remain tightly coupled, making them difficult to optimize jointly under conventional supervised fine-tuning and holistic reinforcement learning paradigms.
To address this challenge, we propose G-CARL, a grounded, checklist-aligned reinforcement learning framework that combines multi-source retrieval for atomic claim verification with context-aware, instance-specific weighted checklists for response coverage, providing structured supervision for factuality, user-demand satisfaction, and expression quality without constraining response diversity.
We further construct \textsc{MMedReport}, a real-world PMRI benchmark, along with a clinician-designed three-dimensional evaluation protocol. Extensive experiments demonstrate that G-CARL consistently outperforms existing post-training baselines in overall quality, claim-level precision, and checklist recall. Pairwise preference evaluation by clinicians further confirms that G-CARL produces interpretations that are more accurate and better aligned with patient needs.
\end{abstract}

\section{Introduction}
Written within a professional medical context, medical reports commonly present abnormal values and descriptive findings without explaining their implications for non-expert readers. 
This gap becomes particularly salient in online healthcare scenarios, where patients upload one or more report images and ask open-ended questions shaped by their personal concerns. 
To bridge this gap, we introduce \emph{Patient-oriented Medical Report Interpretation} (PMRI), a novel open-ended multimodal generation task.
Beyond simply recognizing report findings, PMRI must align professional medical knowledge with patient-centered communication by transforming report evidence into accessible explanations, addressing user-specific concerns, and offering clinically cautious guidance.

\begin{figure}[t]
    \centering
\includegraphics[width=0.99\linewidth]{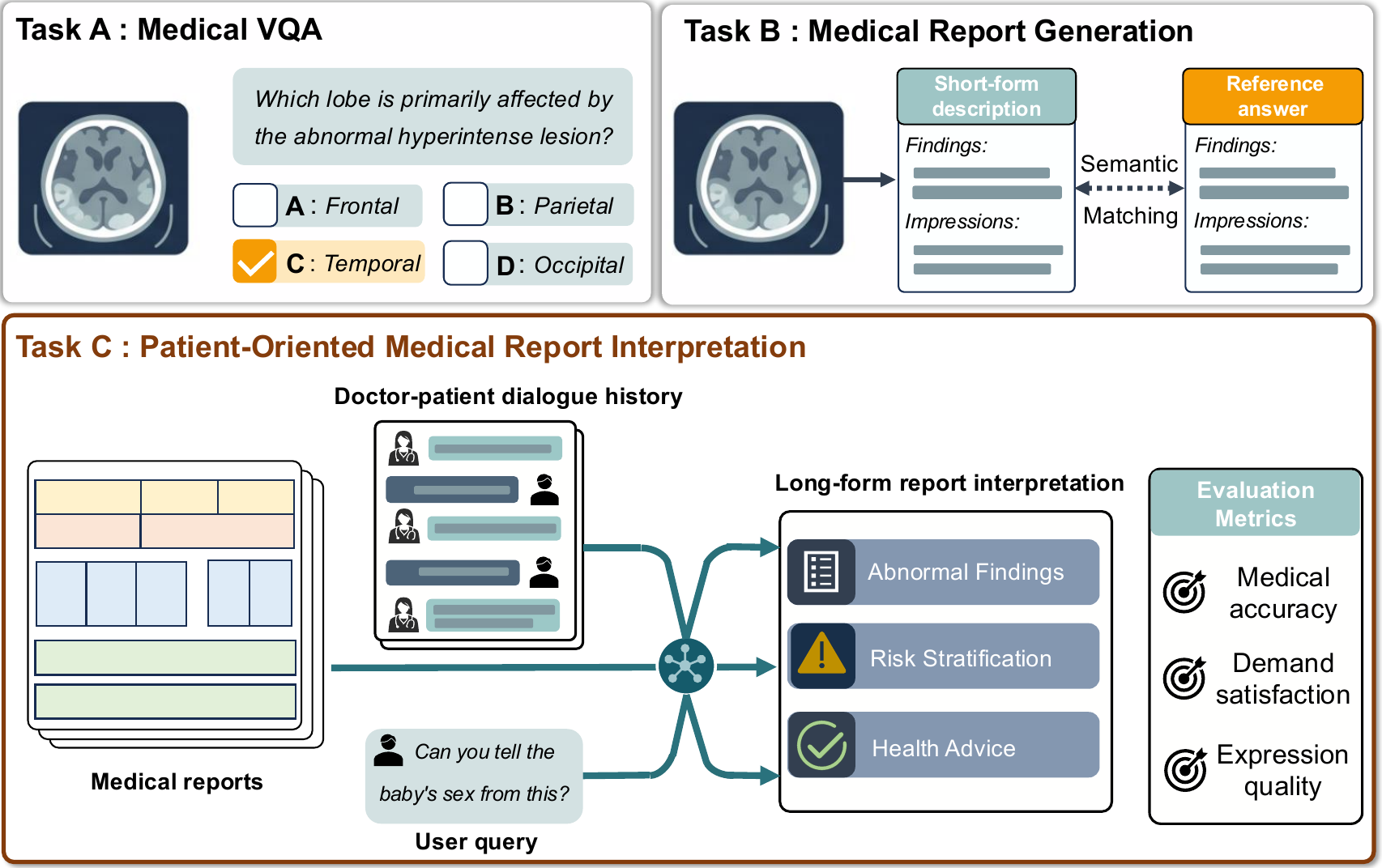}
    \caption{
    Comparison of PMRI (\textcolor{myorange}{\textit{\textbf{Task C}}}) with medical VQA (\textit{\textbf{Task A}}) and conventional report generation (\textit{\textbf{Task B}}). Unlike the short, deterministic outputs of Tasks A and B, PMRI produces long-form, patient-facing explanations grounded in multimodal reports and extended patient–doctor dialogue.
    }
    \vspace{-16pt}
    \label{fig:task_comparison}
\end{figure}

As illustrated in Fig.~\ref{fig:task_comparison}, PMRI differs from conventional medical visual question answering (VQA)~\cite{lu2025bridging,jiang2025knowing} and report generation~\cite{lin2026march,li2025joint} in two important ways. First, it is an \emph{evidence-bounded} generation task whose responses must be grounded not only in the reports but also in reliable clinical knowledge and medically coherent diagnostic reasoning, particularly when explaining abnormalities or suggesting follow-up actions. This requirement is essential for patient safety because unsupported interpretations and inappropriate recommendations may mislead patients or delay necessary care. Second, PMRI is a \emph{patient-facing} communication task that must go beyond clinical correctness to address patients' individual information needs while clearly explaining potential abnormalities and risks in an emotionally adaptive and reassuring manner. Accordingly, we model PMRI quality along three core dimensions: medical accuracy, demand satisfaction, and expression quality.



Given the nature of PMRI, a straightforward strategy is to collect large-scale physician-written interpretations and train multimodal models with supervised fine-tuning (SFT)~\cite{chen2024sharegpt4v,rotstein2024fusecap}. However, SFT can overfit to the specific wording of the reference interpretation and encourage imitation of particular answers rather than learning the underlying principles. This is especially limiting in PMRI, where multiple responses may be clinically acceptable for the same report and user query as long as they remain faithful to the evidence and address the user's concern. Physician references therefore provide useful guidance on what a response should cover, rather than fully defining the quality space of PMRI outputs.

Reinforcement learning (RL) with reward-based post-training, such as Group Relative Policy Optimization (GRPO)~\cite{grpo}, offers a promising alternative for improving large vision-language models beyond reference imitation~\cite{xing2025caprl}. 
However, applying RL to PMRI requires rewards that reflect heterogeneous verifiability of different objectives. Medical factuality can be externally verified against the uploaded report and clinical knowledge, whereas demand satisfaction and expression quality depend more on the patient's concern and the consultation context. This motivates reward signals that can distinguish objective-specific errors rather than collapsing the entire response into a single score.

Existing reward designs do not fully meet this requirement. Holistic MLLM-as-a-Judge scoring~\cite{zheng2023judging,chen2024mllm} compresses an entire interpretation into a single reward, making the supervision coarse and highly dependent on the judge model's medical knowledge. As a result, localized hallucinations may be overlooked when the overall response appears fluent and plausible, despite the fact that a single unsupported medical claim can fundamentally mislead patients. Recent work has introduced rubric-based rewards to provide more structured supervision~\cite{arora2025healthbench,gunjal2025rubrics}. However, static rubrics remain insufficient for PMRI because evaluation priorities vary substantially across cases. 
PMRI errors often stem not from entirely incorrect responses, but from omitting case-critical information or emphasizing secondary details while overlooking the most important clinical recommendations and user concerns. Since static rubrics are designed to capture generic response quality, they are often insensitive to these case-specific omissions and misplaced emphases.


To address these challenges, we propose G-CARL, a reinforcement learning framework with retrieval-grounded and checklist-guided rewards. 
G-CARL assigns reward mechanisms according to the verifiability boundary of each objective. For externally verifiable medical factuality, it decomposes each response into atomic medical claims, retrieves supporting evidence from the uploaded report and a multi-source medical datastore, and evaluates each claim for factual support and contextual relevance. This claim-level reward provides localized supervision for sparse factual errors and discourages unsupported elaboration.
For context-dependent objectives such as demand satisfaction and expression quality, G-CARL constructs instance-specific weighted checklists through MLLM generation followed by clinician refinement. Each checklist item is assigned an automatically generated weight, allowing the reward to emphasize the aspects most relevant to the current report and user question. This yields an explicit checklist score that provides transparent supervision for whether the response addresses the user's concern appropriately.
In summary, our contributions are as follows:
\begin{itemize}

\item We formulate PMRI as an evidence-grounded and patient-facing multimodal generation task, and propose G-CARL, a reinforcement learning framework that decomposes reward supervision according to the heterogeneous verifiability of different objectives.

\item We propose a retrieval-grounded claim reward that provides fine-grained supervision for medical factuality through atomic claim verification, and a case-specific checklist reward that explicitly supervises demand satisfaction and expression quality without relying on a single reference response.

\item We construct \textsc{MMedReport}, a real-world PMRI benchmark with clinician-designed evaluation protocols. Extensive experiments across multiple LVLM backbones demonstrate that G-CARL consistently outperforms supervised and reinforcement learning baselines, with the gains further validated by clinician preference and user comprehension studies.
\end{itemize}








\begin{figure*}[t]
    \centering
    \includegraphics[width=0.95\linewidth]{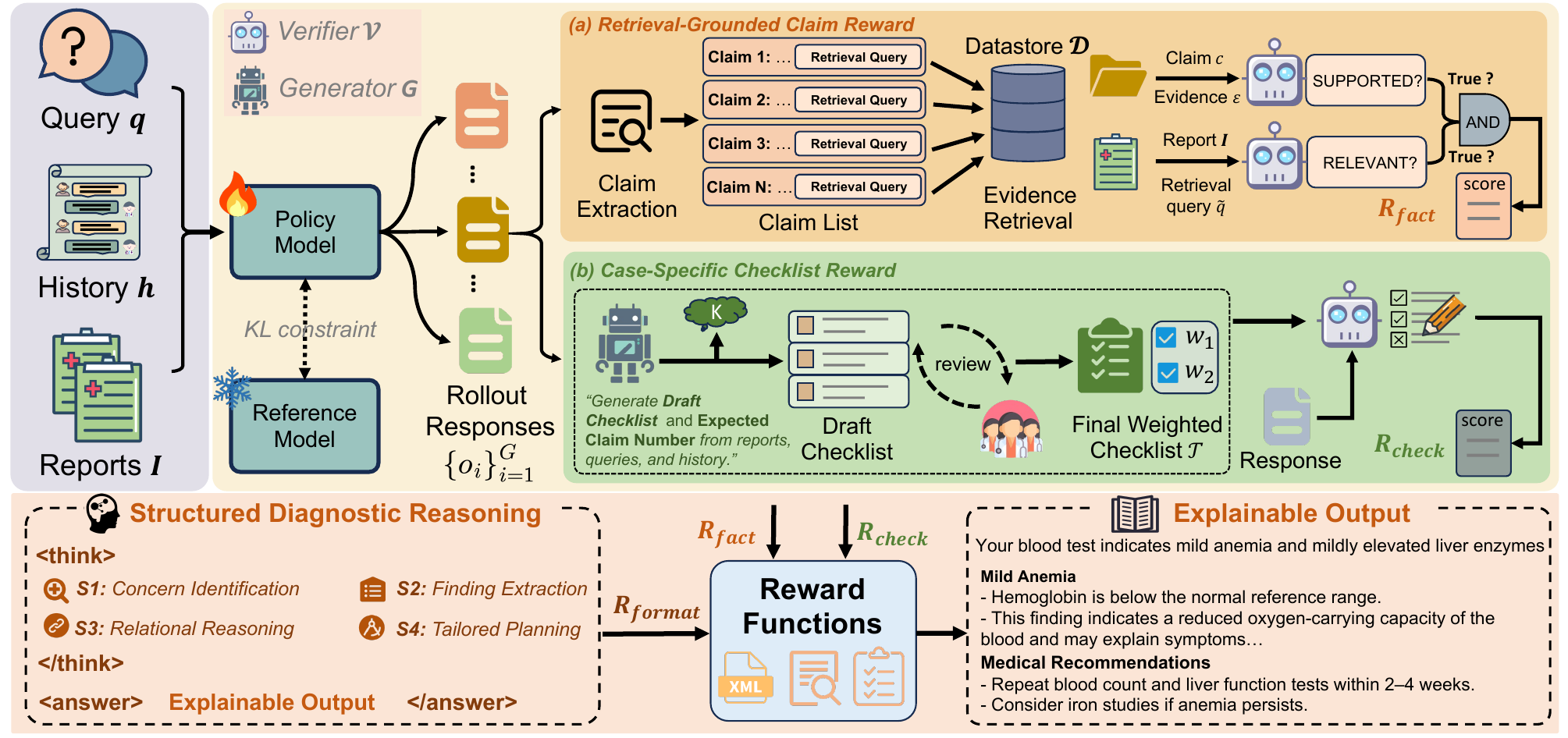}
    \vspace{-8pt}
    \caption{
    Overview of G-CARL. Given report images, user queries, and dialogue history, G-CARL samples $G$ responses from the old policy and optimizes them with a multi-objective reward function consisting of retrieval-grounded claim reward $R_{\mathrm{fact}}$, case-specific checklist reward $R_{\mathrm{check}}$, and structured reasoning format reward $R_{\mathrm{format}}$.
   }
    \label{fig:G-CARL_architecture}
\end{figure*}

\section{Related Works}
\noindent \textbf{Medical Report Generation}. 
Medical report generation has been extensively studied where models generate diagnostic reports from multimodal images such as X-rays~\cite{Li_2023_CVPR,Liu_2025_CVPR}. Recent methods have improved report quality through multimodal feature alignment~\cite{jin2024promptmrg,li2025joint}, clinically grounded visual representations~\cite{varol2025vision}, and reinforcement learning-based optimization~\cite{wang2026beyond}. 
MedVAG~\cite{varol2025vision} introduces clinically aware visual grounding, while HiMed-3B~\cite{wang2026beyond} explores RL-based alignment for medical text generation. MedRepBench~\cite{shang2025medrepbench} further promotes faithful report generation through field-level evaluation of structured clinical findings. In contrast to these imaging-to-report tasks, PMRI focuses on patient-facing interpretation of existing structured reports under patient--doctor dialogue contexts, requiring models to address user-specific concerns while providing accurate and understandable explanations.

\vspace{1pt}
\noindent \textbf{Reinforcement Learning for Medical VLMs}. 
RL has recently been adopted to improve reasoning and reliability in medical VLMs~\cite{jing2026reason,zhou-etal-2026-enhancing}. 
MedVLM-R1~\cite{pan2025medvlm} uses a GRPO-based framework to elicit explicit reasoning paths for radiology VQA, while Med-R1~\cite{lai2026med} designs preference signals that align visual perception, intermediate reasoning, and final answers. 
Beyond short-form QA, MediX-R1~\cite{mullappilly2026medix} extends multimodal medical RL to open-ended responses through LLM-based multi-objective rewards. 
RL has also been explored for report-centric tasks. 
RadVLM-GRPO~\cite{gundersen2026radvlm} applies clinically grounded rewards to chest X-ray report generation and visual grounding, showing that RL can complement strong SFT.

\section{Methods}

\subsection{Overall Architecture}
Built upon GRPO, G-CARL optimizes the policy model with three reward signals, as illustrated in Fig.~\ref{fig:G-CARL_architecture}. Given uploaded medical report images $I$, the dialogue history $h$, and a user query $q$, the policy model first generates candidate responses. The retrieval-grounded branch then supervises medical factuality by verifying report-grounded medical claims against evidence retrieved from a multi-source medical datastore, producing the factuality reward $R_{\mathrm{fact}}$ (\textbf{\textcolor{deepred}{Sec.~\ref{retrieval_grounded}}}). In parallel, the case-specific checklist branch optimizes demand satisfaction and expression quality by constructing a weighted checklist through MLLM generation followed by clinician-guided refinement, and evaluating the checklist coverage of the generated response to produce the checklist reward $R_{\mathrm{check}}$ (\textbf{\textcolor{deepred}{Sec.~\ref{case_specific_check}}}). Together with the format reward $R_{\mathrm{format}}$, these components are integrated into a multi-objective optimization task (\textbf{\textcolor{deepred}{Sec.~\ref{multi_obj_reward}}}).


\subsection{Retrieval-Grounded Claim Reward}
\label{retrieval_grounded}
Medical factuality in PMRI is inherently claim-level, as a single response often contains multiple heterogeneous medical claims. Holistic response-level rewards cannot localize factual errors and may overestimate fluent but unsupported generations. Moreover, many clinical interpretations, causal attributions, and recommendations cannot be verified from the report alone, requiring factual verification grounded in authoritative external medical knowledge. 

To address these challenges, inspired by~\cite{factscore,veriscore}, we propose a retrieval-grounded claim  reward. 
As illustrated in Fig.~\ref{fig:G-CARL_architecture}~(a), each response is first decomposed into atomic medical claims, which are then verified against evidence retrieved from a multi-source medical datastore. 
Specifically, each claim is evaluated from two complementary perspectives: whether it is \emph{RELEVANT} to the uploaded medical report and whether it is \emph{SUPPORTED} by the retrieved evidence.
This dual-binary design encourages the generation of statements that are both clinically grounded and factually correct. We describe each step in detail below.

\vspace{2pt}
\noindent \textbf{Claim Extraction.}
Following~\cite{veriscore}, we perform claim extraction at the sentence level. Given a response $o$, we first segment its answer span into sentences $\{s_1, \dots, s_n\}$. For each sentence $s$, an extractor $\mathcal{G}$, conditioned on the full response $o$ and user query $q$, produces a set of verifiable and decontextualized claims as follows:
\begin{equation}
\{(c, \tilde{q}), \dots\}
=
\mathcal{G}(s \mid o, q).
\end{equation}
Alongside each claim $c$, the extractor jointly generates a retrieval query $\tilde{q}$. 
After claim deduplication, the final claim set for response $o$ is defined as:
\begin{equation}
\mathcal{C}(o)=\{(c_i,\tilde{q}_i)\}_{i=1}^{N}.
\end{equation}

\vspace{2pt}
\noindent \textbf{Evidence Retrieval.}
To support trustworthy evidence-grounded factual verification in the medical domain, we construct a large-scale medical datastore $\mathcal{D}$ by integrating authoritative medical resources, including drug labeling, medical textbooks, and clinical practice guidelines:
\begin{equation}
\mathcal{D} = \mathcal{D}_{\text{drug}} \cup \mathcal{D}_{\text{book}} \cup \mathcal{D}_{\text{guide}},
\end{equation}
where $\mathcal{D}_{\text{drug}}$ contains approximately 20,000 drug instruction entries, $\mathcal{D}_{\text{book}}$ comprises 18,200 textbook passages covering foundational medical knowledge, and $\mathcal{D}_{\text{guide}}$ consists of 16,000 clinical guideline passages. All resources are preprocessed into semantically coherent chunks.

Given a claim-query pair $(c, \tilde{q})$, the retrieval query $\tilde{q}$ combines the topical intent of the user query with the core medical concepts of the corresponding claim, producing a concise set of retrieval keywords aligned with the current clinical context. We then perform parallel retrieval across all knowledge sources using $\tilde{q}$ and aggregate the top-$k$ most relevant evidence chunks from each source:
\begin{equation}
\mathcal{E} = \bigcup \mathrm{Top}_k\big(f( \tilde{q}, \mathcal{D}_i)\big), \quad i \in \{\text{book}, \text{guide}, \text{drug}\},
\end{equation}
where $f(\tilde{q}, d)$ denotes the relevance score between query and document chunk. 
The retrieved evidence set $\mathcal{E}$ is then concatenated with source identifiers and truncated to a fixed token budget to form the evidence context for verifying claim. 

\vspace{2pt}
\noindent \textbf{Dual Binary Verification.}
To verify each extracted claim, a multimodal verifier $\mathcal{V}$ takes the claim $c$, the retrieved evidence $\mathcal{E}$, the user query $q$, and the uploaded report image $I$ as input, and predicts two binary verdicts as:
\begin{equation}
\begin{aligned}
v &= \mathcal{V}(c, \mathcal{E}),
& v &\in \{\textsc{Supported}, \textsc{UnSupported}\},\\
u &= \mathcal{V}(c, I, h, q),
& u &\in \{\textsc{Relevant}, \textsc{Irrelevant}\}, 
\end{aligned}
\end{equation}
where $v$ assesses whether the claim is factually supported, while $u$ assesses whether the claim is contextually relevant to the uploaded reports. 

For factuality verification, the verifier follows an evidence-first, contradiction-oriented protocol. A claim is judged as \textsc{Supported} if it is directly supported by the retrieved evidence or remains consistent with it without contradiction. When explicit evidence is unavailable, the claim is still considered \textsc{Supported} if it reflects established medical consensus and does not conflict with domain knowledge. Otherwise, the claim is judged as \textsc{Unsupported}.


For relevance verification, a claim is judged as \textsc{Relevant} if it is grounded in the uploaded report, directly addresses the user query, or provides clinically necessary context for the current case. Otherwise, it is judged as \textsc{Irrelevant}, even if medically correct, when it introduces generic or unsupported information unrelated to the report findings or the user's concern. This dual-verdict design prevents the verifier from penalizing valid common-sense medical statements that lack explicit retrieved evidence, while discouraging reward hacking through factually correct but irrelevant elaborations. 

Then, a claim is considered valid only if it is both factually supported and contextually relevant: 
\begin{equation}
z
=
\mathbf{1}[v=\textsc{Supported}]
\cdot
\mathbf{1}[u=\textsc{Relevant}]
\in
\{0,1\}.
\end{equation}
The valid-claim precision can be defined as:
\begin{equation}
P_{\mathrm{fact}}
=
\frac{1}{N}
\sum_{i=1}^{N} z_i.
\end{equation}
However, precision alone may favor overly short responses, as a response containing only a few valid claims can still achieve a high precision score. 
To encourage sufficient informational coverage, we introduce a coverage term with a case-specific target claim count $K$, where $K$ is automatically generated by the checklist generator described in Section~\ref{case_specific_check} and represents the expected number of valid medical claims in an informative response:
\begin{equation}
C_{\mathrm{fact}}
=
\min\!\left(
\frac{\sum_{i=1}^{N} z_i}{K}, 
1
\right).
\end{equation}
The final factual validity reward can be formulated as the harmonic mean of valid-claim precision and coverage: 
\begin{equation}
R_{\mathrm{fact}}
=
\frac{2P_{\mathrm{fact}}C_{\mathrm{fact}}}{P_{\mathrm{fact}}+C_{\mathrm{fact}}}.
\end{equation}

\subsection{Case-Specific Checklist Reward}
\label{case_specific_check}
Unlike medical factuality, dimensions such as demand satisfaction and expression quality are inherently subjective and cannot be verified against referenceable knowledge. 
Moreover, these dimensions are highly instance-dependent: 
whether a response is considered complete, helpful, or well-expressed depends on the specific report, user concern, and dialogue context. 
Different criteria also vary substantially in clinical importance. Consequently, assigning a single holistic reward provides little guidance about which aspects of the response should be improved and fails to distinguish critical requirements from desirable but non-essential ones.

To address this challenge, we decompose subjective evaluation into a set of weighted checklist items. Each checklist item represents an independent evaluation criterion with an associated importance weight. A verifier then evaluates the candidate response against each checklist item independently, producing a fine-grained and controllable reward signal for reinforcement learning. The overall procedure is described as follows.

\vspace{2pt}
\noindent \textbf{Weighted Checklist Construction.}
As shown in Fig.~\ref{fig:G-CARL_architecture}~(b), to reduce manual annotation effort, an MLLM (\textit{e.g.}, Gemini 3.1 Pro) as generator $G$ first generates a draft instance-specific checklist conditioned on the dialogue history, user query, and uploaded medical reports. Professional clinicians then iteratively review, refine, and complete the draft to produce the final checklist:
\begin{equation}
\mathcal{T}
=
\{(t_i, w_i)\}_{i=1}^{m},
\end{equation}
where $t_i$ denotes a self-contained checklist item and $w_i$ its corresponding importance weight. 
Physician-guided checklist construction grounds the supervision signal in clinically appropriate expectations rather than generic evaluation templates, which prior work has shown to be essential for reliable expert-domain assessment~\cite{viswanathan2025checklists}.

The importance weight $w$ is determined according to the clinical significance of each checklist item. Specifically, each checklist item is assigned one of four importance levels, namely \textsc{Essential}, \textsc{Important}, \textsc{Optional}, or \textsc{Pitfall}, which are mapped to predefined integer-valued weights (e.g., 4--5, 2--3, 1--2, and $-2$--$-1$, respectively). This weighting scheme ensures that clinically critical criteria contribute more strongly to the reward while undesirable behaviors incur explicit penalties.
In particular, \textsc{Pitfall} items capture undesirable behaviors discouraged by the physicians, including ignoring the user's primary concern, generating overly technical explanations, or providing dismissive responses.

\vspace{2pt}
\noindent \textbf{Explicit Aggregation.}
Given the checklist $\mathcal{T}$ and a candidate response $o$, the checklist verifier $\mathcal{V}$ evaluates each checklist item independently and outputs a binary satisfaction signal:
\begin{equation}
z_i = \mathcal{V}(t_i, o) \in \{0,1\}.
\end{equation}
For \textsc{Essential}, \textsc{Important}, and \textsc{Optional} items, $z_i=1$ indicates that the criterion is satisfied. For \textsc{Pitfall} items, $z_i=1$ indicates that the undesirable behavior is triggered. 
We denote the positive checklist item set as $\mathcal{P}$ and the \textsc{Pitfall} item set as $\mathcal{N}$, and compute the positive coverage and pitfall violation terms as: 
\begin{equation}
V_{\mathrm{pos}}
=
\frac{
\sum_{i\in\mathcal{P}} |w_i| z_i
}{
\sum_{i\in\mathcal{P}} |w_i|
},
\qquad
V_{\mathrm{neg}}
=
\frac{
\sum_{j\in\mathcal{N}} |w_j| z_j
}{
\sum_{i\in\mathcal{P}} |w_i|
}.
\end{equation}
The final checklist reward is defined as:
\begin{equation}
R_{\mathrm{check}}
=
\mathrm{clip}\!\left(
V_{\mathrm{pos}} - V_{\mathrm{neg}},
0,
1
\right).
\end{equation}
Unlike conventional reference-based training, our checklist does not encourage the policy to mimic a single physician-authored response. Instead, the reference is distilled into a set of clinically important evaluation criteria, specifying \emph{what} the response should cover rather than \emph{how} it should be written. This design preserves the diversity of valid responses while rewarding clinical completeness and user-oriented communication.

\subsection{Multi-objective Reward Function}
\label{multi_obj_reward}
Inspired by recent advances in structured reasoning~\cite{yu2025docthinker}, we introduce a format reward that encourages the model to follow a physician-oriented clinical reasoning scaffold during reinforcement learning. As illustrated in Fig.~\ref{fig:G-CARL_architecture}, the model is required to organize its intermediate reasoning within the \texttt{<think>}\texttt{</think>} block according to a four-step clinical reasoning workflow before generating the final patient-facing interpretation in the \texttt{<answer>}\texttt{</answer>} block. This reward does not directly supervise medical correctness; instead, it encourages a structured decomposition of report findings, supporting evidence, clinical reasoning, and response planning, thereby improving the coherence and organization of the final response.
Our reward function jointly optimizes medical factuality, demand satisfaction and expression quality, which can be defined as:
\begin{equation}
R_\text{total}
=
\lambda_{\text{fact}}R_\text{fact}
+
\lambda_{\text{check}}R_\text{check}
+
\lambda_{\text{format}}R_\text{format},
\end{equation}
where $\lambda_{\mathrm{fact}}$, $\lambda_{\mathrm{check}}$, and $\lambda_{\mathrm{format}}$ are weighting coefficients that balance the contributions of the reward components.


\section{Experiments}

\subsection{Experimental Setup}
\noindent \textbf{Datasets.} 
We evaluate G-CARL primarily on \textbf{(\textcolor{deepred}{1})} \textsc{MMedReport}, a real-world multimodal PMRI benchmark collected from online healthcare consultations. It contains 2,450 instances, each including dialogue history, a user query, uploaded medical report images, and clinician-verified reference annotations. All instances undergo quality control, de-identification, and manual verification to ensure data quality and patient privacy. Detailed dataset statistics are provided in \textbf{Appendix~A}. Following the standard split, 2,200 instances are used for training and 250 for evaluation. 
To assess broader medical capability beyond PMRI, we further evaluate the trained models on an external benchmark, \textbf{(\textcolor{deepred}{2})} \textsc{CMB}~\cite{wang2024cmb}, which evaluates both medical QA accuracy and the professionalism of open-ended clinical interpretation under an LLM-as-a-judge protocol.


\vspace{1pt}
\noindent \textbf{Evaluation Protocol.} We evaluate PMRI responses using both subjective and objective metrics. (1) \emph{Subjective Evaluation}.
Responses are evaluated along three clinician-defined dimensions: \emph{Medical Accuracy} (Accuracy), \emph{Demand Satisfaction} (Satisfaction), and \emph{Expression Quality} (Expression). Each dimension is scored according to a detailed clinician-authored rubric ranging from $-2$ to $3$. We employ GPT-5.2~\cite{openai2025gpt52systemcard} as the judge by strictly following these evaluation criteria. 
The complete judging protocol is provided in the \textbf{Appendix B} to facilitate reproducibility.
(2) \emph{Objective Evaluation}.
In addition to holistic assessment, we report claim-level \emph{Precision} and checklist-level \emph{Recall}. Precision is defined as the ratio of supported claims to extracted claims, measuring factual correctness, while Recall is defined as the ratio of satisfied checklist items to the total checklist items, measuring coverage of case-specific requirements. 

\vspace{1pt}
\noindent \textbf{Implementation Details.}
Our policy models are initialized from the Qwen3-VL~\cite{bai2025qwen3vltechnicalreport} and InternVL3~\cite{zhu2025internvl3exploringadvancedtraining} series. G-CARL is trained for three epochs on 8 NVIDIA H100 GPUs with a batch size of 192 and a learning rate of $1\times10^{-5}$. 
We set the number of rollout responses to $G=8$ during training. 
The reward weights are configured as $\lambda_{\text{fact}}=0.4$, $\lambda_{\text{check}}=0.3$, and $\lambda_{\text{format}}=0.3$; detailed hyperparameter analysis is provided in \textbf{Appendix C}. 
The verifier $\mathcal{V}$ is initialized with Qwen3.5-35B-A3B, with implementation details reported in \textbf{Appendix D}. 
We compare G-CARL with the corresponding base models (Base), supervised fine-tuning (SFT), and an MLLM-as-a-Judge reward baseline. 
Additionally, we evaluate a diverse range of general-purpose and medical LVLMs under a zero-shot inference setting.

\begin{table*}[!t]
\centering
\caption{Main results on our medical report interpretation benchmark. Numbers are mean over 3 seeds with standard deviation. }
\vspace{-8pt}
\label{tab:main}
\setlength{\tabcolsep}{9.2pt}
\renewcommand{\arraystretch}{0.92}
\scalebox{0.85}{
\begin{tabular}{l|cccc|cc}
\toprule
\multirow{2}{*}{\textbf{Method}}
& \multicolumn{4}{c|}{\textbf{Subjective Metrics}}
& \multicolumn{2}{c}{\textbf{Objective Metrics}} \\
\cmidrule(lr){2-5}\cmidrule(lr){6-7}
& \textbf{Overall} & Accuracy & Satisfaction & Expression & Precision (\%) & Recall (\%) \\
\midrule
\rowcolor{rowgray}
\multicolumn{7}{l}{\textit{General LVLMs}} \\
GPT-4o~\cite{hurst2024gpt}          
& 1.588$_{\pm0.007}$ & 0.974$_{\pm0.007}$ & 0.358$_{\pm0.001}$ & 0.256$_{\pm0.001}$ & 95.26 & 42.63 \\
ERNIE 4.5 VL~\cite{ernie45technicalreport}          
& 1.709$_{\pm0.010}$ & 1.088$_{\pm0.012}$ & 0.376$_{\pm0.002}$ & 0.245$_{\pm0.000}$ & 95.27 & 53.74 \\
GLM-4.6V~\cite{glm46v}              
& 1.819$_{\pm0.014}$ & 1.159$_{\pm0.013}$ & 0.395$_{\pm0.002}$ & 0.265$_{\pm0.001}$ & 95.90 & 59.51 \\
Step-3.7-Flash~\cite{step37flashblog} 
& 1.842$_{\pm0.011}$ & 1.175$_{\pm0.012}$ & 0.412$_{\pm0.002}$ & 0.255$_{\pm0.001}$ & 95.96 & 72.11 \\
Kimi K2.5~\cite{kimi2.5}            
& 1.903$_{\pm0.002}$ & 1.221$_{\pm0.002}$ & 0.418$_{\pm0.001}$ & 0.264$_{\pm0.001}$ & 97.63 & 76.42 \\
\rowcolor{rowours}
Gemini 3.1 Pro~\cite{gemini31pro}
& \textbf{1.964$_{\pm0.008}$} & \textbf{1.229$_{\pm0.008}$} & \textbf{0.438$_{\pm0.003}$} & \textbf{0.297$_{\pm0.002}$} & \textbf{98.10} & \textbf{80.29} \\
\midrule
\rowcolor{rowgray}
\multicolumn{7}{l}{\textit{Medical LVLMs}} \\
Hulumed~\cite{hulumed}         
& 1.020$_{\pm0.018}$ & 0.480$_{\pm0.013}$ & 0.282$_{\pm0.004}$ & 0.258$_{\pm0.002}$ & 72.28 & 32.65 \\
Medgemma~\cite{sellergren2026medgemma} 
& 1.004$_{\pm0.005}$ & 0.460$_{\pm0.008}$ & 0.289$_{\pm0.004}$ & 0.255$_{\pm0.001}$ & 74.30 & 33.86 \\
Lingshu~\cite{lingshu}
& \textbf{1.527$_{\pm0.003}$} & \textbf{0.902$_{\pm0.002}$} & \textbf{0.357$_{\pm0.003}$} & \textbf{0.268$_{\pm0.002}$} & \textbf{89.69} & \textbf{39.41} \\
\midrule
\rowcolor{rowgray}
\multicolumn{7}{l}{\textit{Qwen-VL series}} \\
Qwen3-VL-4B-Instruct (Base)
& 1.527$_{\pm0.010}$ & 0.897$_{\pm0.007}$ & 0.374$_{\pm0.003}$ & 0.256$_{\pm0.001}$ & 92.19 & 49.35 \\
\quad +SFT
& 1.603$_{\pm0.009}$ & 0.945$_{\pm0.012}$ & 0.392$_{\pm0.002}$ & 0.266$_{\pm0.002}$ & 92.90 & 57.28 \\
\quad +MLLM-as-a-Judge
& 1.662$_{\pm0.012}$ & 0.993$_{\pm0.009}$ & 0.396$_{\pm0.002}$ & 0.273$_{\pm0.001}$ & 93.82 & 57.72 \\
\rowcolor{rowpink}
\quad \textbf{+Ours}
& \textbf{1.709$_{\pm0.009}$} & \textbf{1.028$_{\pm0.008}$} & \textbf{0.406$_{\pm0.000}$} & \textbf{0.275$_{\pm0.002}$} & \textbf{93.97} & \textbf{58.92} \\
Qwen3-VL-8B-Instruct (Base)
& 1.626$_{\pm0.012}$ & 0.980$_{\pm0.010}$ & 0.388$_{\pm0.006}$ & 0.258$_{\pm0.001}$ & 93.46 & 60.68 \\
\quad +SFT
& 1.718$_{\pm0.008}$ & 1.040$_{\pm0.010}$ & 0.403$_{\pm0.002}$ & 0.275$_{\pm0.001}$ & 94.22 & 63.44 \\
\quad +MLLM-as-a-Judge
& 1.766$_{\pm0.008}$ & 1.089$_{\pm0.007}$ & 0.407$_{\pm0.002}$ & 0.270$_{\pm0.001}$ & 95.85 & 65.47 \\
\rowcolor{rowpink}
\quad \textbf{+Ours}
& \textbf{1.829$_{\pm0.009}$} & \textbf{1.141$_{\pm0.008}$} & \textbf{0.411$_{\pm0.001}$} & \textbf{0.277$_{\pm0.001}$} & \textbf{96.62} & \textbf{72.18} \\
\midrule
\rowcolor{rowgray}
\multicolumn{7}{l}{\textit{InternVL3 series}} \\
InternVL3-8B (Base)
& 1.358$_{\pm0.007}$ & 0.744$_{\pm0.007}$ & 0.345$_{\pm0.001}$ & 0.269$_{\pm0.001}$ & 91.25 & 40.37 \\
\quad +SFT
& 1.513$_{\pm0.018}$ & 0.868$_{\pm0.016}$ & 0.385$_{\pm0.004}$ & 0.260$_{\pm0.001}$ & 92.06 & 50.88 \\
\quad +MLLM-as-a-Judge   
& 1.553$_{\pm0.018}$ & 0.897$_{\pm0.017}$ & 0.388$_{\pm0.001}$ & 0.271$_{\pm0.001}$ & 92.39 & 52.61 \\
\rowcolor{rowpink}
\quad \textbf{+Ours}         
& \textbf{1.638$_{\pm0.008}$} & \textbf{0.967$_{\pm0.010}$} & \textbf{0.397$_{\pm0.001}$} & \textbf{0.274$_{\pm0.001}$} & \textbf{92.51} & \textbf{57.97} \\
InternVL3-14B (Base)
& 1.616$_{\pm0.016}$ & 0.986$_{\pm0.013}$ & 0.365$_{\pm0.004}$ & 0.265$_{\pm0.001}$ & 92.93 & 44.86 \\
\quad +SFT
& 1.663$_{\pm0.007}$ & 0.995$_{\pm0.008}$ & 0.400$_{\pm0.000}$ & 0.268$_{\pm0.001}$ & 93.34 & 58.58 \\
\quad +MLLM-as-a-Judge          
& 1.670$_{\pm0.003}$ & 1.010$_{\pm0.004}$ & 0.399$_{\pm0.003}$ & 0.261$_{\pm0.000}$ & 93.72 & 58.06 \\
\rowcolor{rowpink}
\quad \textbf{+Ours}         
& \textbf{1.739$_{\pm0.007}$}
& \textbf{1.061$_{\pm0.005}$}
& \textbf{0.405$_{\pm0.000}$}
& \textbf{0.273$_{\pm0.002}$}
& \textbf{95.42}
& \textbf{64.57} \\
\bottomrule
\end{tabular}}
\end{table*}



\begin{table}[t]
\centering
\setlength{\tabcolsep}{6 pt}
\renewcommand{\arraystretch}{0.9}
\caption{Evaluation results on the CMB dataset.}
\vspace{-8pt}
\label{tab:cmb_evaluation}
\scalebox{0.83}{
\begin{tabular}{lccc}
\toprule[1.5pt]
\multirow{2}{*}{Method}
& \multicolumn{2}{c}{QA Accuracy (\%)}
& \multicolumn{1}{c}{Open-ended Generation} \\
\cmidrule(lr){2-3}\cmidrule(lr){4-4}
& Train & Val & Professionalism \\
\midrule
Base            & 74.85 & 71.42 & 3.58 \\
SFT             & 75.00 & 71.43 & 3.48 \\
MLLM-as-a-Judge & 74.96 & 70.36 & 3.51 \\
\textbf{G-CARL} & \textbf{75.48} & \textbf{72.05} & \textbf{3.61} \\
\bottomrule[1.5pt]
\end{tabular}}
\vspace{-6pt}
\end{table}


\begin{table}[t]
\centering
\renewcommand{\arraystretch}{0.85}
\caption{Ablation study of reward designs in G-CARL.}
\vspace{-8pt}
\label{tab:tab2}
\setlength{\tabcolsep}{2.8pt}
\scalebox{0.83}{
\begin{tabular}{lcccccc}
\toprule[1.5pt]
    & \textbf{Overall} & Acc & Sat & Exp & Pre & Rec \\
\midrule
\multicolumn{7}{l}{Reward Combination} \\
\midrule
Baseline
    & 1.626 & 0.980 & 0.388 & 0.258 & 93.46 & 60.68 \\
+ $R_\text{check}$ + $R_\text{format}$
    & 1.720 & 1.046 & 0.406 & 0.268 & 95.36 & 64.10 \\
+ $R_\text{fact}$ + $R_\text{format}$
    & 1.748 & 1.074 &0.404 & 0.270 & 95.84 & 62.81 \\
+ $R_\text{fact}$ + $R_\text{check}$
    & 1.810 & 1.135 & 0.411 & 0.264 & 96.03 & 70.09 \\
+ $R_\text{fact}$ + $R_\text{check}$ + $R_\text{format}$
    & \textbf{1.829} & \textbf{1.141} & \textbf{0.411}
    & \textbf{0.277} & \textbf{96.62} & \textbf{72.18} \\
\midrule
\multicolumn{7}{l}{Reward Design}     \\
\midrule
$R_\text{check}$: w/ static rubric
    & 1.786 & 1.100 & 0.412 & 0.274 & 94.83 & 66.09 \\
$R_\text{fact}$: w/o retrieval
    & 1.734 & 1.057 & 0.406 & 0.271 & 94.26 & 63.81      \\
\bottomrule[1.5pt]
\end{tabular}}
\vspace{-18pt}
\end{table}

\subsection{Main Results on MMedReport}
\noindent \textbf{Quantitative Comparison.}
Table~\ref{tab:main} compares G-CARL with general-purpose LVLMs, specialized medical LVLMs, and different training paradigms based on the Qwen3-VL and InternVL3 backbones. 
While the MLLM-as-a-Judge reward improves over SFT, its holistic rubric struggles to jointly optimize medical accuracy, demand satisfaction, and expression quality. 
By decomposing reward supervision into externally verifiable medical claims and internally grounded checklist objectives, G-CARL achieves the highest scores across both objective and subjective metrics. 
On Qwen3-VL-8B, G-CARL improves the overall subjective score, while boosting claim-level precision (+0.77\%) and checklist-level recall (+6.71\%), 
indicating more informative and richer interpretations. 
Moreover, G-CARL consistently outperforms specialized medical LVLMs and remains competitive with substantially larger general-purpose LVLMs under zero-shot evaluation, demonstrating the effectiveness of G-CARL.

\begin{figure*}[t]
    \centering
\includegraphics[width=0.95\linewidth]{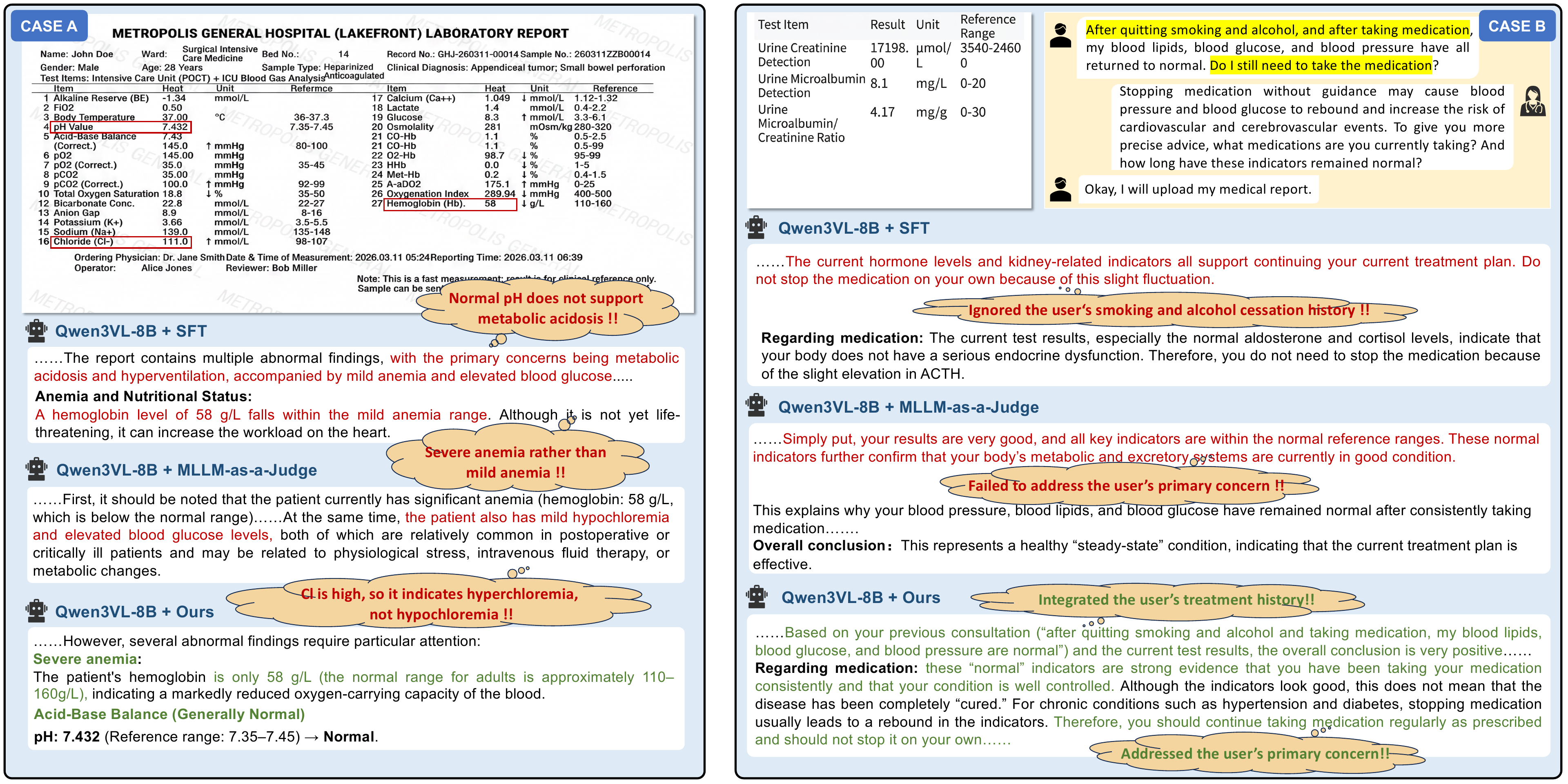}
\vspace{-8pt}
    \caption{Qualitative comparison of interpretation results across different models. Two representative real-world cases (Case A and Case B) are presented, where Case B includes both report images and dialogue history as input.}
    \vspace{-10pt}
    \label{fig:visualize}
\end{figure*}

\vspace{1pt}
\noindent \textbf{External Generalization Study.}
We further transfer the trained models to CMB without any adaptation. As shown in Table~\ref{tab:cmb_evaluation}, G-CARL improves QA accuracy on both splits (+0.63) and attains the highest professionalism score in open-ended generation (3.61), whereas SFT and MLLM-as-a-Judge bring marginal gains and even degrade professionalism. This indicates that grounding rewards in verifiable medical evidence suppresses hallucinated content and better elicits the medical accuracy already latent in the base model, rather than merely fitting the PMRI response style.

\vspace{1pt}
\noindent \textbf{Case Study.}
As shown in Fig. \ref{fig:visualize}, we qualitatively compare the outputs of our method with SFT and MLLM-as-a-Judge using Qwen3VL-8B as the base model. The report shows a normal pH (7.432), elevated chloride (111.0 mmol/L), and severe anemia (Hb = 58 g/L). However, SFT incorrectly diagnoses metabolic acidosis and misclassifies the severe anemia as mild, while MLLM-as-a-Judge mistakes the elevated chloride level for hypochloremia instead of hyperchloremia. In contrast, our method correctly identifies all key abnormalities and produces clinically accurate interpretations. Similarly, in the right example, SFT fails to incorporate the user's smoking and alcohol cessation history, while MLLM-as-a-Judge fails to address the user's primary concern of whether to continue the medication. 
\begin{figure}[h]
    \centering    \includegraphics[width=0.99\linewidth]{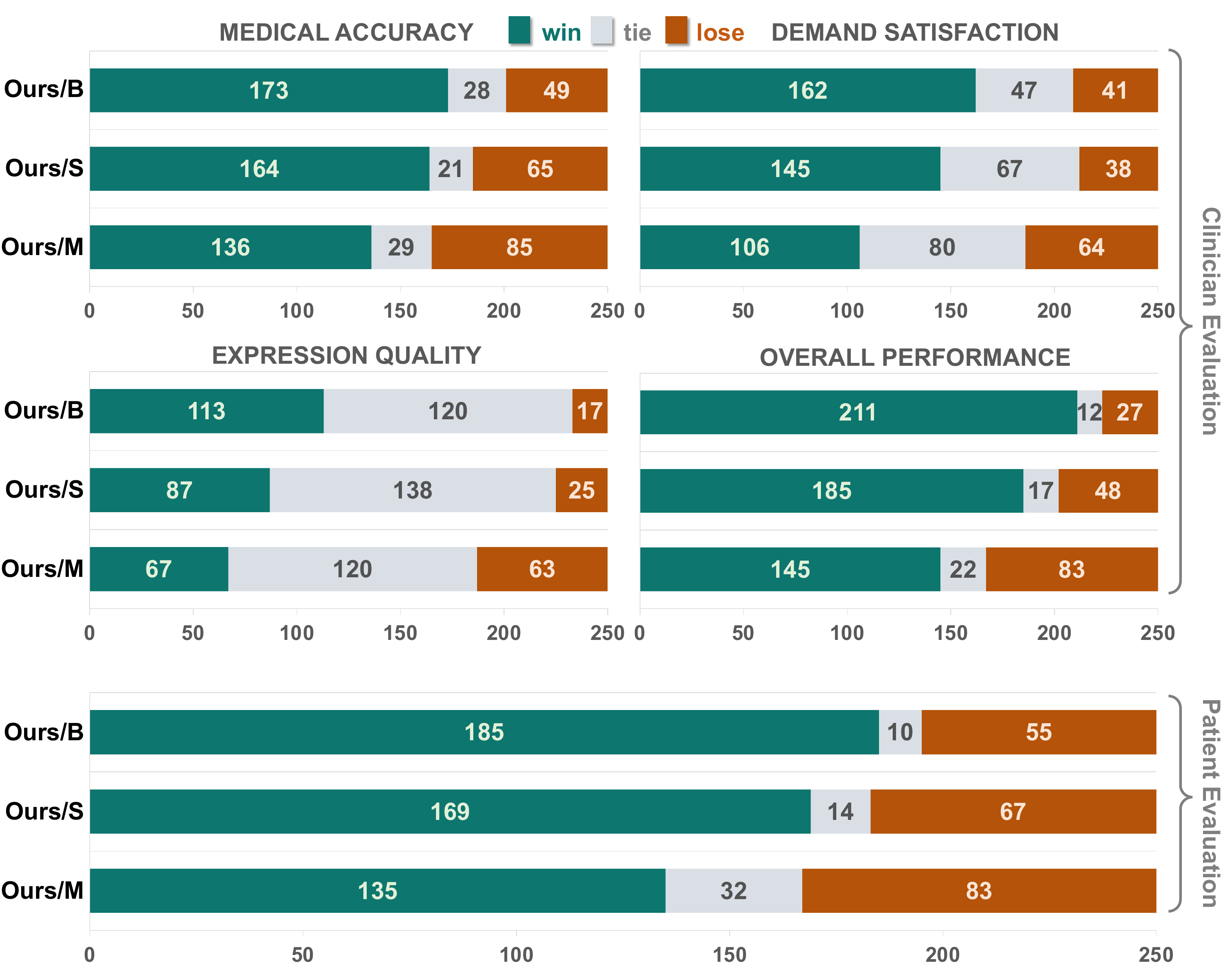}
    \vspace{-8pt}
    \caption{Human preference evaluation by clinicians and patients. "Ours/B", "Ours/S", and "Ours/M" denote pairwise comparisons between G-CARL and the base model, the SFT model, and the MLLM-as-a-Judge-trained model.}
    \vspace{-18pt}
    \label{fig:preference}
\end{figure}
In contrast, our method effectively integrates historical context, directly answers the user's question, and provides evidence-grounded recommendations.


\noindent \textbf{Human Preference Evaluation.}
We conduct a blind pairwise human preference study on 250 held-out cases. Three professional clinicians compare anonymized responses generated by G-CARL and the baselines across the three evaluation dimensions and overall preference, with majority voting used for the final decision. We additionally recruit 50 participants without medical training to evaluate the comprehensibility of the generated interpretations.
As shown in Fig.~\ref{fig:preference}, G-CARL is consistently preferred over both SFT and MLLM-as-a-Judge, particularly in medical accuracy and demand satisfaction, with preference margins of 136:85 and 106:64, respectively. It also achieves substantially higher comprehensibility ratings from non-expert participants, suggesting that its responses are easier for patients to understand.


\subsection{Ablation Study}
\noindent \textbf{Effect of Reward Components.} 
To evaluate the contribution of each reward component, we conduct ablation studies on Qwen3VL-8B. 
As shown in Table~\ref{tab:tab2}, $R_\text{check}$ combined with $R_\text{format}$ improves the baseline by enhancing case-specific requirement coverage, while $R_\text{fact}$ provides substantial gains in medical factuality. 
Combining all three rewards achieves the best performance, demonstrating their complementary effects. 
Further analysis shows that replacing the dynamic checklist with a static rubric degrades overall performance, highlighting the importance of case-adaptive supervision. 
Similarly, removing retrieval from $R_\text{fact}$ reduces claim-level precision, confirming the effectiveness of retrieval-grounded factual verification.

\vspace{1pt}
\noindent \textbf{Comparison with Other RL Methods.}
Table~\ref{tab:tab3} compares G-CARL with representative RL methods for open-ended generation. Preference-based DPO performs worst, since high-quality preference data can hardly cover the diverse response space of PMRI. PROMETHEUS scores responses directly against reference answers, which is too coarse-grained to yield confident judgments and thus provides unstable reward signals. Rubric-based RAR offers finer criteria but lacks explicit evidence verification, while MedRepBench emphasizes structured clinical finding recall and is therefore less aligned with user-specific demands. Factuality-oriented rewards (FactScore, CapRL) bring the most pronounced accuracy gains among the baselines. By coupling retrieval-grounded claim verification with objective-specific reward decomposition, G-CARL attains the best overall performance, jointly improving three dimensions.

\begin{table}[t]
\centering
\setlength{\tabcolsep}{2.5pt}
\renewcommand{\arraystretch}{0.92}
\caption{Comparison with existing RL methods.}
\vspace{-10pt}
\label{tab:tab3}
\scalebox{0.83}{
\begin{tabular}{l|cccc}
\toprule[1.5pt]
Method & \textbf{Overall} & Acc & Sat & Exp \\
\midrule
DPO~\cite{rafailov2023direct} & 1.032 & 0.438 & 0.332 & 0.262 \\
\midrule
\multicolumn{5}{l}{GRPO-based methods} \\
\midrule
\quad + PROMETHEUS~\cite{kim2024prometheus} & 1.132 & 0.530 & 0.347 & 0.255 \\
\quad + RAR~\cite{gunjal2025rubrics} & 1.683 & 1.016 & 0.398 & 0.269 \\
\quad + MedRepBench~\cite{shang2025medrepbench} & 1.722 & 1.048 & 0.406 & 0.268 \\
\quad + CapRL~\cite{xing2025caprl} & 1.729 & 1.050 & 0.410 & 0.269 \\
\quad + FactScore~\cite{min2023factscore} & 1.776 & 1.100 & 0.408 & 0.268 \\
\quad + \textbf{Ours} & \textbf{1.829} & \textbf{1.141} & \textbf{0.411} & \textbf{0.277} \\
\bottomrule[1.5pt]
\end{tabular}}
\vspace{-14pt}
\end{table}

\section{Conclusion}
In this paper, we present PMRI as a challenging yet underexplored task that requires both evidence-grounded medical factuality and context-dependent patient communication. We propose G-CARL, a reinforcement learning framework that decomposes reward supervision through retrieval-grounded claim verification and case-specific checklist guidance. Extensive experiments on the newly constructed \textsc{MMedReport} benchmark, together with clinician-authored evaluation and human preference studies, show that G-CARL consistently improves the quality of patient-oriented medical report interpretation. We hope this work lays the foundation for more reliable and patient-centered multimodal medical assistants.


\bibliography{aaai2027}

@inproceedings{factscore,
  title={Factscore: Fine-grained atomic evaluation of factual precision in long form text generation},
  author={Min, Sewon and Krishna, Kalpesh and Lyu, Xinxi and Lewis, Mike and Yih, Wen-tau and Koh, Pang and Iyyer, Mohit and Zettlemoyer, Luke and Hajishirzi, Hannaneh},
  booktitle={Proceedings of the 2023 Conference on Empirical Methods in Natural Language Processing},
  pages={12076--12100},
  year={2023}
}

@inproceedings{veriscore,
  author       = {Yixiao Song and others},
  title        = {VeriScore: Evaluating the factuality of verifiable claims in long-form
                  text generation},
  pages        = {9447--9474},
  publisher    = {Association for Computational Linguistics},
  year         = {2024}
}

@inproceedings{wang2026beyond,
  title={Beyond n-grams: A hierarchical reward learning framework for clinically-aware medical report generation},
  author={Wang, Yuan and Gao, Shujian and Liu, Jiaxiang and Jiang, Songtao and Haoxiang, Xia and Zhang, Xiaotian and Kang, Zhaolu and Wang, Yemin and Liu, Zuozhu},
  booktitle={Proceedings of the AAAI Conference on Artificial Intelligence},
  volume={40},
  number={40},
  pages={33719--33727},
  year={2026}
}

@article{varol2025vision,
  title={A vision attention driven Language framework for medical report generation},
  author={Varol Arisoy and others},
  journal={Scientific Reports},
  volume={15},
  number={1},
  pages={10704},
  year={2025},
  publisher={Nature Publishing Group UK London}
}

@article{shang2025medrepbench,
  title={Medrepbench: A comprehensive benchmark for medical report interpretation},
  author={Shang, Fangxin and Xia, Yuan and Yang, Dalu and Wang, Yahui and Yang, Binglin},
  journal={arXiv preprint arXiv:2508.16674},
  year={2025}
}

@inproceedings{jin2024promptmrg,
  title={Promptmrg: Diagnosis-driven prompts for medical report generation},
  author={Jin, Haibo and Che, Haoxuan and Lin, Yi and Chen, Hao},
  booktitle={Proceedings of the AAAI conference on artificial intelligence},
  volume={38},
  number={3},
  pages={2607--2615},
  year={2024}
}

@article{li2025joint,
  title={Joint Imbalance Adaptation for Radiology Report Generation},
  author={Li, Wang and Han, Guangzeng and Wu, Yuexin and Huang, I-Chan and Huang, Xiaolei},
  journal={Journal of Healthcare Informatics Research},
  pages={1--23},
  year={2025},
  publisher={Springer}
}

@inproceedings{pan2025medvlm,
  title={Medvlm-r1: Incentivizing medical reasoning capability of vision-language models (vlms) via reinforcement learning},
  author={Pan, Jiazhen and Liu, Che and Wu, Junde and Liu, Fenglin and Zhu, Jiayuan and Li, Hongwei Bran and Chen, Chen and Ouyang, Cheng and Rueckert, Daniel},
  booktitle={International Conference on Medical Image Computing and Computer-Assisted Intervention},
  pages={337--347},
  year={2025},
  organization={Springer}
}

@article{lai2026med,
  title={Med-r1: Reinforcement learning for generalizable medical reasoning in vision-language models},
  author={Lai, Yuxiang and Zhong, Jike and Li, Ming and Zhao, Shitian and Li, Yuheng and Psounis, Konstantinos and Yang, Xiaofeng},
  journal={IEEE transactions on medical imaging},
  year={2026},
  publisher={IEEE}
}

@article{lin2026march,
  title={MARCH: Multi-Agent Radiology Clinical Hierarchy for CT Report Generation},
  author={Lin, Yi and Ding, Yihao and Wu, Yonghui and Peng, Yifan},
  journal={arXiv preprint arXiv:2604.16175},
  year={2026}
}

@article{xing2025caprl,
  title={Caprl: Stimulating dense image caption capabilities via reinforcement learning},
  author={Xing, Long and Dong, Xiaoyi and Zang, Yuhang and Cao, Yuhang and Liang, Jianze and Huang, Qidong and Wang, Jiaqi and Wu, Feng and Lin, Dahua},
  journal={arXiv preprint arXiv:2509.22647},
  year={2025}
}

@inproceedings{chen2024mllm,
  title={MLLM-as-a-Judge: Assessing Multimodal LLM-as-a-Judge with Vision-Language Benchmark},
  author={Chen, Dongping and Chen, Ruoxi and Zhang, Shilin and Liu, Yinuo and Wang, Yaochen and Zhou, Huichi and Zhang, Qihui and Zhou, Pan and Wan, Yao and Sun, Lichao},
  booktitle={International Conference on Machine Learning},
  year={2024}
}

@article{gundersen2026radvlm,
  title={RadVLM-GRPO: Enhancing Chest X-ray Report Generation and Visual Grounding via Reinforcement Learning},
  author={Gundersen, Benjamin and Deperrois, Nicolas and Ruiperez-Campillo, Samuel and Sutter, Thomas M and Vogt, Julia E and Moor, Michael and Nooralahzadeh, Farhad and Krauthammer, Michael},
  journal={Proceedings of Machine Learning Research},
  volume={150},
  pages={1--34},
  year={2026}
}

@article{mullappilly2026medix,
  title={Medix-r1: Open ended medical reinforcement learning},
  author={Mullappilly, Sahal Shaji and Kurpath, Mohammed Irfan and Mohamed, Omair and Zidan, Mohamed and Khan, Fahad and Khan, Salman and Anwer, Rao and Cholakkal, Hisham},
  journal={arXiv preprint arXiv:2602.23363},
  year={2026}
}

@inproceedings{viswanathan2025checklists,
  title     = {Checklists Are Better Than Reward Models For Aligning Language Models},
  author    = {Viswanathan, Vijay and Sun, Yanchao and Ma, Shuang and Kong, Xiang and Cao, Meng and Neubig, Graham and Wu, Tongshuang},
  booktitle = {Advances in Neural Information Processing Systems},
  year      = {2026}
}

@misc{openai2025gpt52systemcard,
  title        = {Update to GPT-5 System Card: GPT-5.2},
  author       = {{OpenAI}},
  howpublished = {\url{https://cdn.openai.com/pdf/3a4153c8-c748-4b71-8e31-aecbde944f8d/oai_5_2_system-card.pdf}},
  year         = {2025},
  month        = dec
}

@article{hurst2024gpt,
  title={Gpt-4o system card},
  author={Hurst, Aaron and Lerer, Adam and Goucher, Adam P and Perelman, Adam and Ramesh, Aditya and Clark, Aidan and Ostrow, AJ and Welihinda, Akila and Hayes, Alan and Radford, Alec and others},
  journal={arXiv preprint arXiv:2410.21276},
  year={2024}
}

@misc{kimi2.5,
      title={Kimi K2.5: Visual Agentic Intelligence}, 
      author={Kimi Team and Tongtong Bai and Yifan Bai and Yiping Bao and others},
      year={2026},
      eprint={2602.02276},
      archivePrefix={arXiv},
      primaryClass={cs.CL},
      url={https://arxiv.org/abs/2602.02276}, 
}

@misc{glm46v,
      title={GLM-4.5V and GLM-4.1V-Thinking: Towards Versatile Multimodal Reasoning with Scalable Reinforcement Learning}, 
      author={V Team and others},
      year={2026},
      eprint={2507.01006},
      archivePrefix={arXiv},
      primaryClass={cs.CV},
      url={https://arxiv.org/abs/2507.01006}, 
}

@article{lingshu,
  title={Lingshu: A generalist foundation model for unified multimodal medical understanding and reasoning},
  author={Xu, Weiwen and Chan, Hou Pong and Li, Long and Aljunied, Mahani and Yuan, Ruifeng and Wang, Jianyu and Xiao, Chenghao and Chen, Guizhen and Liu, Chaoqun and Li, Zhaodonghui and others},
  journal={arXiv preprint arXiv:2506.07044},
  year={2025}
}

@misc{gemini31pro,
  title        = {Gemini 3.1 Pro Model Card},
  author       = {{Google DeepMind}},
  year         = {2026},
  month        = feb,
  howpublished = {\url{https://deepmind.google/models/model-cards/gemini-3-1-pro}},
  note         = {Published February 2026}
}

@misc{ernie45technicalreport,
  title        = {ERNIE 4.5 Technical Report},
  author       = {{Baidu ERNIE Team}},
  year         = {2025},
  howpublished = {\url{https://ernie.baidu.com/blog/publication/ERNIE_Technical_Report.pdf}},
  note         = {Technical report}
}

@misc{step37flashblog,
  title        = {Step 3.7 Flash: A High-Efficiency Flash Model for Real-World Agentic Workflows},
  author       = {{StepFun AI}},
  year         = {2026},
  month        = may,
  howpublished = {\url{https://static.stepfun.com/blog/step-3.7-flash}},
  note         = {Official blog post}
}

@article{hulumed,
  title={Hulu-med: A transparent generalist model towards holistic medical vision-language understanding},
  author={Jiang, Songtao and Wang, Yuan and Song, Sibo and Hu, Tianxiang and Zhou, Chenyi and Pu, Bin and Zhang, Yan and Yang, Zhibo and Feng, Yang and Zhou, Joey Tianyi and others},
  journal={arXiv preprint arXiv:2510.08668},
  year={2025}
}

@article{sellergren2026medgemma,
  title={Medgemma 1.5 technical report},
  author={Sellergren, Andrew and Gao, Chufan and Mahvar, Fereshteh and Kohlberger, Timo and Jamil, Fayaz and Traverse, Madeleine and Tono, Alberto and Sadjad, Bashir and Yang, Lin and Lau, Charles and others},
  journal={arXiv preprint arXiv:2604.05081},
  year={2026}
}

@inproceedings{kim2024prometheus,
  title={Prometheus: Inducing fine-grained evaluation capability in language models},
  author={Kim, Seungone and Shin, Jay and Jang, Joel and Longpre, Shayne and Lee, Hwaran and Yun, Sangdoo and Shin, Ryan and Kim, Sungdong and Thorne, James and Seo, Minjoon and others},
  booktitle={International Conference on Learning Representations},
  volume={2024},
  pages={29927--29962},
  year={2024}
}

@article{rafailov2023direct,
  title={Direct preference optimization: Your language model is secretly a reward model},
  author={Rafailov, Rafael and Sharma, Archit and Mitchell, Eric and Manning, Christopher D and Ermon, Stefano and Finn, Chelsea},
  journal={Advances in neural information processing systems},
  volume={36},
  pages={53728--53741},
  year={2023}
}

@inproceedings{min2023factscore,
  title={Factscore: Fine-grained atomic evaluation of factual precision in long form text generation},
  author={Min, Sewon and Krishna, Kalpesh and Lyu, Xinxi and Lewis, Mike and Yih, Wen-tau and Koh, Pang and Iyyer, Mohit and Zettlemoyer, Luke and Hajishirzi, Hannaneh},
  booktitle={Proceedings of the 2023 Conference on Empirical Methods in Natural Language Processing},
  pages={12076--12100},
  year={2023}
}

@article{gunjal2025rubrics,
  title={Rubrics as Rewards: Reinforcement Learning Beyond Verifiable Domains},
  author={Gunjal, Anisha and Wang, Anshul Vijay and Yao, Chi and Chen, Jinghan and Lo, Kalinda and Rane, Nikhil and Yang, Sara and Zheng, Yangming and Wang, Zhengfei and others},
  journal={arXiv preprint arXiv:2507.17746},
  year={2025}
}

@article{lu2025bridging,
  title={Bridging the Semantic Gap in Medical Visual Question Answering With Prompt Learning},
  author={Lu, Zilin and Zeng, Qingjie and Lu, Mengkang and Chen, Geng and Xia, Yong},
  journal={IEEE Transactions on Medical Imaging},
  volume={44},
  number={11},
  pages={4605--4616},
  year={2025},
  doi={10.1109/TMI.2025.3580561}
}

@inproceedings{jiang2025knowing,
  title={Knowing or Guessing? Robust Medical Visual Question Answering via Joint Consistency and Contrastive Learning},
  author={Jiang, Songtao and Chen, Yuxi and Song, Sibo and Zhang, Yan and Jin, Yeying and Feng, Yang and Wu, Jian and Liu, Zuozhu},
  booktitle={International Conference on Medical Image Computing and Computer-Assisted Intervention},
  pages={325--335},
  year={2025},
  organization={Springer}
}

@inproceedings{chen2024sharegpt4v,
  title={Sharegpt4v: Improving large multi-modal models with better captions},
  author={Chen, Lin and Li, Jinsong and Dong, Xiaoyi and Zhang, Pan and He, Conghui and Wang, Jiaqi and Zhao, Feng and Lin, Dahua},
  booktitle={European Conference on Computer Vision},
  pages={370--387},
  year={2024},
  organization={Springer}
}

@inproceedings{rotstein2024fusecap,
  title={Fusecap: Leveraging large language models for enriched fused image captions},
  author={Rotstein, Noam and Bensaid, David and Brody, Shaked and Ganz, Roy and Kimmel, Ron},
  booktitle={Proceedings of the IEEE/CVF winter conference on applications of computer vision},
  pages={5689--5700},
  year={2024}
}

@inproceedings{zheng2023judging,
  title={Judging LLM-as-a-Judge with MT-Bench and Chatbot Arena},
  author={Zheng, Lianmin and Chiang, Wei-Lin and Sheng, Ying and Zhuang, Siyuan and Wu, Zhanghao and Zhuang, Yonghao and Lin, Zi and Li, Zhuohan and Li, Dacheng and Xing, Eric P. and Zhang, Hao and Gonzalez, Joseph E. and Stoica, Ion},
  booktitle={Advances in Neural Information Processing Systems},
  year={2023}
}

@article{arora2025healthbench,
  title={HealthBench: Evaluating Large Language Models Towards Improved Human Health},
  author={Arora, Rahul K. and Wei, Jason and Soskin Hicks, Rebecca and Bowman, Preston and Qui\~{n}onero-Candela, Joaquin and Tsimpourlas, Foivos and Sharman, Michael and Shah, Meghan and Vallone, Andrea and Beutel, Alex and Heidecke, Johannes and Singhal, Karan},
  journal={arXiv preprint arXiv:2505.08775},
  year={2025}
}

@inproceedings{yu2025docthinker,
  title={Docthinker: Explainable multimodal large language models with rule-based reinforcement learning for document understanding},
  author={Yu, Wenwen and Yang, Zhibo and Liu, Yuliang and Bai, Xiang},
  booktitle={Proceedings of the IEEE/CVF International Conference on Computer Vision},
  pages={837--847},
  year={2025}
}

@inproceedings{wang2024cmb,
  title={CMB: A Comprehensive Medical Benchmark in Chinese},
  author={Wang, Xidong and Chen, Guiming Hardy and Song, Dingjie and Zhang, Zhiyi and Chen, Zhihong and Xiao, Qingying and Jiang, Feng and Li, Jianquan and Wan, Xiang and Wang, Benyou and Li, Haizhou},
  booktitle={Proceedings of the 2024 Conference of the North American Chapter of the Association for Computational Linguistics: Human Language Technologies},
  year={2024},
  note={arXiv:2308.08833}
}

@misc{bai2025qwen3vltechnicalreport,
      title={Qwen3-VL Technical Report}, 
      author={Shuai Bai and others},
      year={2025},
      eprint={2511.21631},
      archivePrefix={arXiv},
      primaryClass={cs.CV},
      url={https://arxiv.org/abs/2511.21631}, 
}

@misc{zhu2025internvl3exploringadvancedtraining,
      title={InternVL3: Exploring Advanced Training and Test-Time Recipes for Open-Source Multimodal Models}, 
      author={Jinguo Zhu and others},
      year={2025},
      eprint={2504.10479},
      archivePrefix={arXiv},
      primaryClass={cs.CV},
      url={https://arxiv.org/abs/2504.10479}, 
}

@article{jing2026reason,
  title = {Reason like a radiologist: Chain-of-thought and reinforcement learning for verifiable report generation},
  journal = {Medical Image Analysis},
  volume = {109},
  pages = {103910},
  year = {2026},
  issn = {1361-8415},
  doi = {https://doi.org/10.1016/j.media.2025.103910},
  author = {Peiyuan Jing and Kinhei Lee and Zhenxuan Zhang and Huichi Zhou and Zhengqing Yuan and Zhifan Gao and Lei Zhu and Giorgos Papanastasiou and Yingying Fang and Guang Yang},
}

@inproceedings{zhou-etal-2026-enhancing,
    title = "Enhancing Reinforcement Learning for Radiology Report Generation with Evidence-aware Rewards and Self-correcting Preference Learning",
    author = "Zhou, Qin and
      Liang, Guoyan and
      Yang, Qianyi and
      Chen, Jingyuan and
      Wu, Sai and
      Yao, Chang and
      Wang, Zhe",
    year = "2026",
    booktitle = "Proceedings of the 64th Annual Meeting of the Association for Computational Linguistics (Volume 1: Long Papers)",
    publisher = "Association for Computational Linguistics",
    pages = "37044--37056",
    isbn = "979-8-89176-390-6"
}

@InProceedings{Li_2023_CVPR,
    author    = {Li, Mingjie and Lin, Bingqian and Chen, Zicong and Lin, Haokun and Liang, Xiaodan and Chang, Xiaojun},
    title     = {Dynamic Graph Enhanced Contrastive Learning for Chest X-Ray Report Generation},
    booktitle = {Proceedings of the IEEE/CVF Conference on Computer Vision and Pattern Recognition (CVPR)},
    month     = {June},
    year      = {2023},
    pages     = {3334-3343}
}

@InProceedings{Liu_2025_CVPR,
    author    = {Liu, Kang and Ma, Zhuoqi and Kang, Xiaolu and Li, Yunan and Xie, Kun and Jiao, Zhicheng and Miao, Qiguang},
    title     = {Enhanced Contrastive Learning with Multi-view Longitudinal Data for Chest X-ray Report Generation},
    booktitle = {Proceedings of the IEEE/CVF Conference on Computer Vision and Pattern Recognition (CVPR)},
    month     = {June},
    year      = {2025},
    pages     = {10348-10359}
}

@article{grpo,
  title={Deepseekmath: Pushing the limits of mathematical reasoning in open language models},
  author={Shao, Zhihong and Wang, Peiyi and Zhu, Qihao and Xu, Runxin and Song, Junxiao and Bi, Xiao and Zhang, Haowei and Zhang, Mingchuan and Li, YK and Wu, Yang and others},
  journal={arXiv preprint arXiv:2402.03300},
  year={2024}
}


\end{document}